\documentclass[10pt,twocolumn,letterpaper]{article}

\usepackage{cvpr}              

\usepackage{graphicx}
\usepackage{amsmath}
\usepackage{amssymb}
\usepackage{booktabs}
\usepackage[accsupp]{axessibility}

\usepackage[pagebackref,breaklinks,colorlinks]{hyperref}

\usepackage[capitalize]{cleveref}
\crefname{section}{Sec.}{Secs.}
\Crefname{section}{Section}{Sections}
\Crefname{table}{Table}{Tables}
\crefname{table}{Tab.}{Tabs.}

\def\cvprPaperID{5} 
\def\confName{CVPR}
\def\confYear{2023}

\begin{document}

\title{ODSmoothGrad: Generating Saliency Maps for Object Detectors}

\author{Chul Gwon\\
Analytic Folk\\
{\tt\small chul@analyticfolk.com}
\and
Steven C. Howell\\
ARLIS \\
University of Maryland\\
College Park, MD\\
{\tt\small showell@arlis.umd.edu}
}
\maketitle

\begin{abstract}
	Techniques for generating saliency maps continue to be used for explainability of deep learning models, with efforts primarily applied to the image classification task. Such techniques, however, can also be applied to object detectors, not only with the classification scores, but also for the bounding box parameters, which are regressed values for which the relevant pixels contributing to these parameters can be identified. In this paper, we present ODSmoothGrad, a tool for generating saliency maps for the classification and the bounding box parameters in object detectors. Given the noisiness of saliency maps, we also apply the SmoothGrad algorithm \cite{smilkov2018} to visually enhance the pixels of interest. We demonstrate these capabilities on one-stage and two-stage object detectors, with comparisons using classifier-based techniques.
\end{abstract}

\begin{figure}[t]
  \centering
  \begin{subfigure}{\linewidth}
    \includegraphics[width=0.9\linewidth]{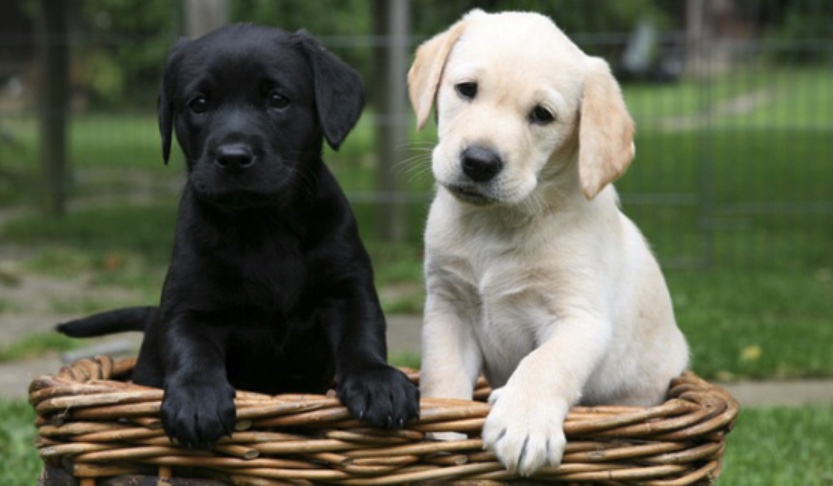}
    \caption{Original Image}
  \end{subfigure}
  \begin{subfigure}{\linewidth}
    \includegraphics[width=0.9\linewidth]{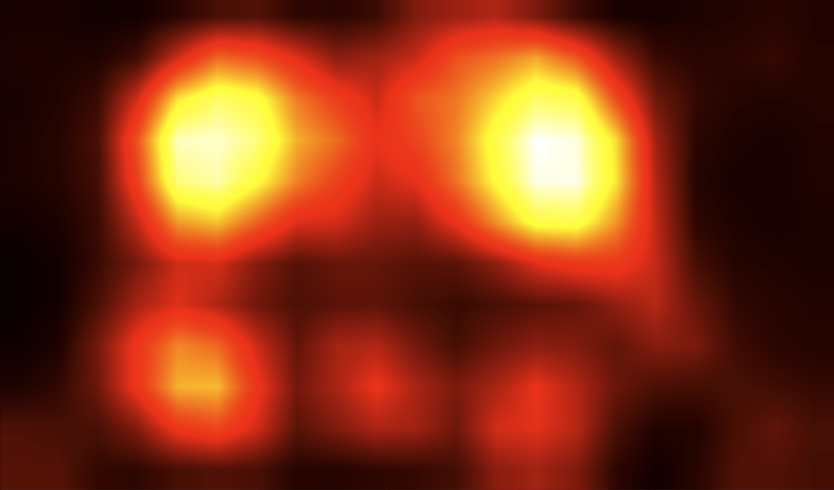}
    \caption{Grad-CAM}
  \end{subfigure}
  \begin{subfigure}{\linewidth}
    \includegraphics[width=0.9\linewidth]{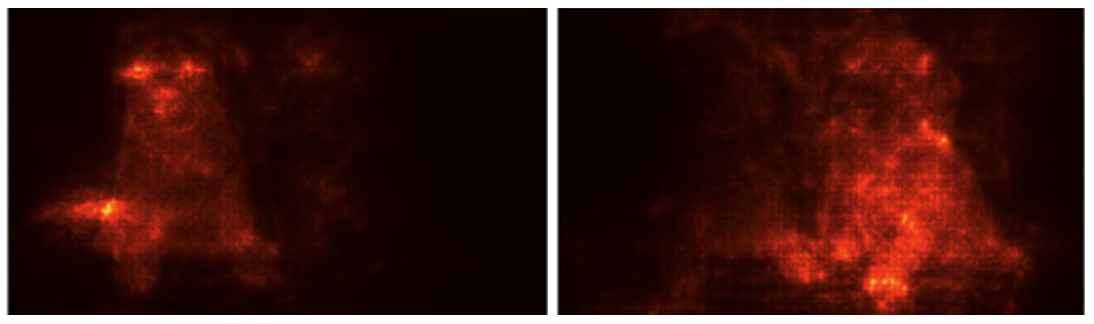}
    \caption{ODSmoothGrad}
  \end{subfigure}
  \caption{Saliency maps for classification and object detection for an image with two Labrador retrievers shown in (a). Classification-based maps generate a single map for the Labrador retriever class - the Grad-CAM mask using ResNet-101 is shown in (b). Object detection-based maps generate separate masks for each detected object rather than a single mask - ODSmoothGrad maps for the classification output of Faster R-CNN shown in (c).}
  \label{fig:saliency}
\end{figure}

\begin{figure*}
  \centering
  \includegraphics[width=0.9\linewidth]{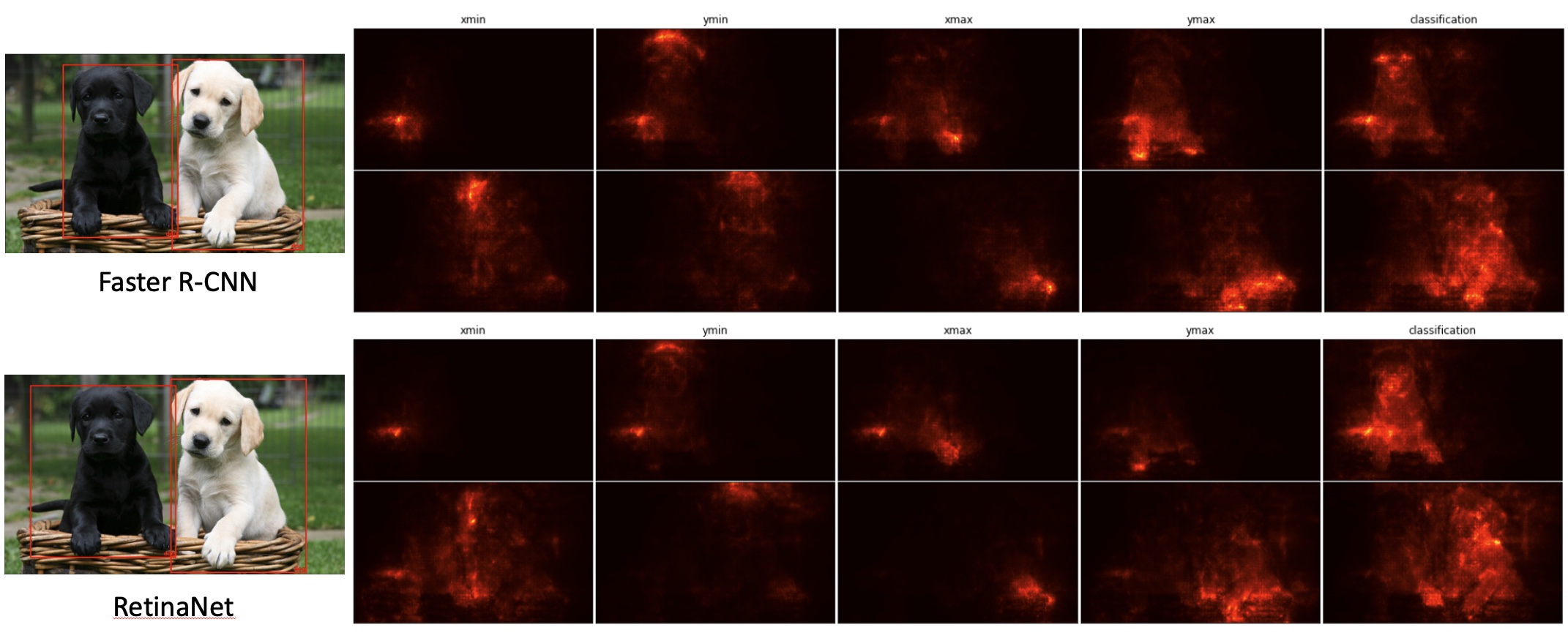}
  \caption{ODSmoothGrad saliency maps generated for Faster R-CNN and RetinaNet from the Detectron2 Model Zoo. The top row of each shows the saliency maps for the dog on the left, while the bottom row shows the dog on the right. The columns show the saliency map for $x_{min}$ (left), $y_{min}$ (top), $x_{max}$ (right), $y_{max}$ (bottom), and classification.}
  \label{fig:odsg}
\end{figure*}

\section{Introduction}
\label{sec:intro}

There is a significant amount of work on algorithms and tools for improving the explainability of models. As models become more complex, it becomes increasingly difficult to determine how the model achieved a particular result. Methods for obtaining saliency maps have long been used to highlight the parts of an image that provide the greatest contribution to a given output. The majority of the effort has been on the image classification problem, with some growing interest in the field of object detectors \cite{petsiuk2020}, although these approaches still focus on explainability of the classification results and ignore the bounding box values.

Extending the use of saliency maps to gain visibility into the decision-making process of object detectors is also relevant for explainability into these types of models. The class of a detected object, along with the bounding box parameters that localize it within the image, are all derived from the model. Confirming that these returned values are based on relevant features from the image is important for verifying model performance. The bounding box parameters become particularly important when there are shifts in the predicted box as opposed to the ground truth annotation, or if ground truth annotations have intentionally been sized slightly larger to include some context. Another motivation for saliency maps with object detectors is that when there are multiple objects of the same class in a given image, the object detector will return a separate map for each detected object, while the classifier will return a single map for the entire image (Figure \ref{fig:saliency}).

The contributions made by this paper include the application of saliency methods to object detectors to identify relevant pixels for both the classification and bounding box parameters, as well as demonstrating this capability on one-stage (RetinaNet \cite{lin2018}) and two-stage object detectors (Faster RCNN \cite{ren2015}). Our implementation applies the SmoothGrad algorithm \cite{smilkov2018} to improve visibility of relevant pixels.

\section{Related Work}
\label{sec:related_work}

There is significant work in the field of object detection, including transformer-based object detectors \cite{carion2020}. For this study, we focused our work on anchor-based object detectors, particularly RetinaNet \cite{lin2018} and Faster R-CNN \cite{ren2015}, to demonstrate the one-stage and two-stage detectors. Although these are no longer state-of-the-art for object detection benchmarks, the ability to train these models with more modest-sized datasets and hardware still make them popular options for detection.

As mentioned previously, the majority of the work in this area has been done for the image classification problem. Along with saliency-based approaches, recent work extended Layer-wise Relevance Propagation (LRP) specifically for use with SSDs \cite{tsunakawa2019} and YOLO5 \cite{karasmanoglu2022}. The techniques applied here focus solely on the classification importance, rather than also including the relevance of the bounding box parameters themselves.

\begin{figure*}
  \centering
  \begin{subfigure}{0.9\linewidth}
    \includegraphics[width=\linewidth]{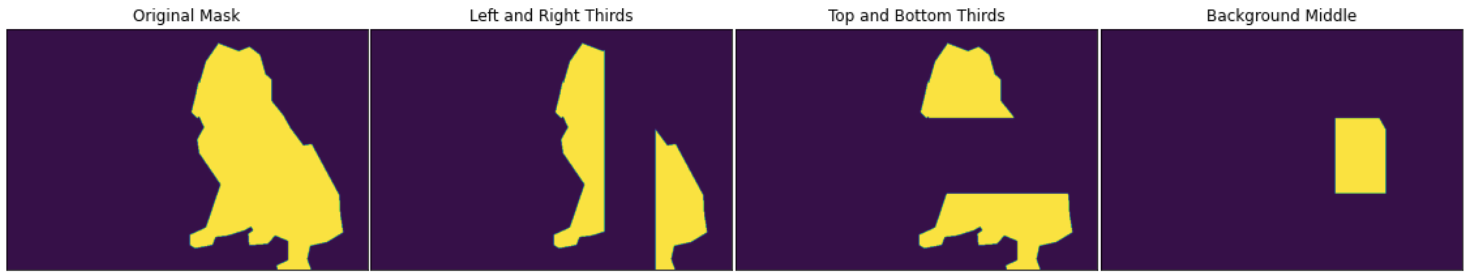}
    \caption{Original ground truth mask divided into thirds for validation}
    \label{fig:val_a}
  \end{subfigure}
  \begin{subfigure}{0.9\linewidth}
    \includegraphics[width=\linewidth]{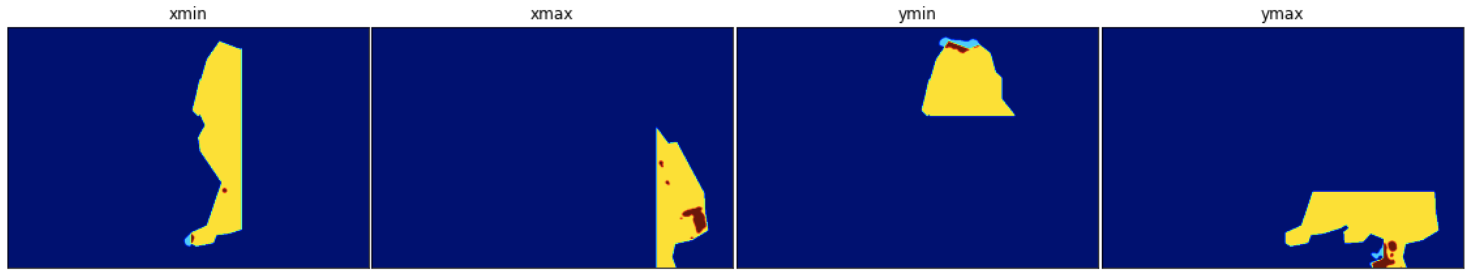}
    \caption{Overlay of ground truth and prediction masks for the bounding box components}
    \label{fig:val_b}
  \end{subfigure}
  \caption{To validate the performance of the masks for the bounding box parameters, the original ground truth mask was divided into thirds, with the middle third used for background comparisons (a). The intersection-over-foreground (IOF) was then calculated between the saliency masks and the corresponding third of the ground truth mask (b). Original mask in yellow, saliency mask in cyan, and overlap region in dark red.}
  \label{fig:val}
\end{figure*}

\subsection{Saliency Methods with Classification}

There are several published techniques that discuss the use of saliency methods, along with one that goes over the problems that particular saliency methods exhibit \cite{adebayo2018}. We provide a quick overview of some of them here, especially those relevant to our work and related work, but this is not intended to be an exhaustive list.

Class Saliency Extraction \cite{simonyan2014} begins with an image, with $m$ rows and $n$ columns, from which a saliency map $M \in R^{m\times n}$ is generated in the following manner: perform backward pass from the logit of interest to obtain derivative $w$; take the absolute value of each element of $M$; for color im- ages, take the max value across the channels for each pixel.

SmoothGrad \cite{smilkov2018} takes the average of multiple samples of class saliency extraction, injecting random noise into the image for each sample. The resulting map calculation looks as follows:

\begin{equation}
	M_c(x) = \frac{1}{n} \sum_{1}^{n} M_c(x+\mathcal{N}(0,\sigma^2))
\end{equation}

\noindent where $M \in R^{m\times n}$, $n$ is the number of samples, and $\mathcal{N}(0,\sigma^2)$ is Gaussian-distributed noise with a standard deviation $\sigma$.

Grad-CAM \cite{selvaraju2016} performs a backward pass from the logit of particular class to the last convolutional layer in the CNN, and then average pools across the width and height dimensions to produce a weight value for each of the channels in the final layer with respect to this class. These weights are multiplied by the corresponding activation map, and the maps are combined and passed through a ReLU operation to create the final map. Since there are multiple downsampling operations that occur from the input image to the final convolutional layer, the width and height of the final layer is less than the initial image. To create a map of the same dimension as the original image, a bilinear upsampling is used, which still results in a coarse map. To achieve a pixel-level saliency map, Grad-CAM is combined with Guided Backpropagation to create Guided Grad-CAM.

\subsection{Saliency Methods with Object Detection}

For applying saliency methods to object detection, there have been three examples that have done similar work to ours, albeit using different techniques. The first is DetGrad-CAM \cite{saha2019}, which applies Grad-CAM for all of the features and then sums across these features to produce a final saliency map. In this manner, DetGrad-CAM does allow for localized saliency maps, as expected from object detectors, but their work was restricted to classification improvements, and they tested it only on YOLOv2.

The second method is Spatial Sensitive Grad-CAM (SSGradCAM) \cite{yamauchi2022}, which applies Grad-CAM to SSD. This technique only uses the classification output from the detected object, but does not take the bounding box parameters into consideration. It was also developed for use with the original SSD \cite{ssd2016}.

The third method takes a different approach than the gradient-based approaches discussed so far. D-RISE \cite{petsiuk2020} generates a series of masks for a given image, and passes the image with the different masks applied through the object detector. The class probabilities generated by these masked images are used as the importance weights, and the masks are then combined based on these weights to produce the final importance map.

\section{Method}

We performed our experiment using two methods, the first being more true to the saliency extraction calculation from the logit, and the second using a simplified implementation that uses SmoothGrad and could be quickly generalized to other architectures. 

For the first implementation, using RetinaNet \cite{lin2017}, we found the anchor boxes with classification scores beyond a particular threshold. For each of these, we performed the backward pass against the classification logit as well as each of the bounding box parameters (xmin, ymin, xmax, ymax) to generate five saliency maps. Using this technique, we would not only be able to determine the saliency map for a particular class, but also the relevant pixels for other classes, as is possible with saliency map creation. The major problem with this implementation was that it was specific for RetinaNet, and so generalizing this for use across different object detectors would not be as straightforward.

For the second method, rather than going through the anchor boxes directly, we took the output tensors generated from the outputs from Detectron2 \cite{wu2019} and ran the backward pass from those directly. Although this goes against starting the backward pass from the logit, when combined with SmoothGrad, the resulting saliency maps were clearer than with a single pass from our first implementation. For the SmoothGrad step, we chose a sample size $n$ = 20, and a noise value $\sigma$ = 0.05.

Between each of the sampling passes for SmoothGrad, we needed to align the bounding boxes from the detections. To accomplish this, we applied a threshold of 0.7 on the value of the intersection-over-union (IOU) between the bounding boxes of detected objects between each pass. For each detected object, we took averages of each of the saliency maps generated for the classification and bounding box parameters to produce the final results.

\section{Experiments}
\subsection{Implementation}

Whereas ODSmoothGrad could be applied against general object detector implementations, we specifically developed our library to work with Detectron2. We used RetinaNet and Faster RCNN from the Detectron2 Model Zoo, using implementations with the ResNet-101 backbone and a version of Faster RCNN that included the Feature Pyramid Network \cite{lin2018}. As shown in Figure \ref{fig:odsg}, for each detected object, saliency maps are generated for the four bounding box parameters and the classification label.

\begin{figure}[t]
  \centering
  \includegraphics[width=0.9\linewidth]{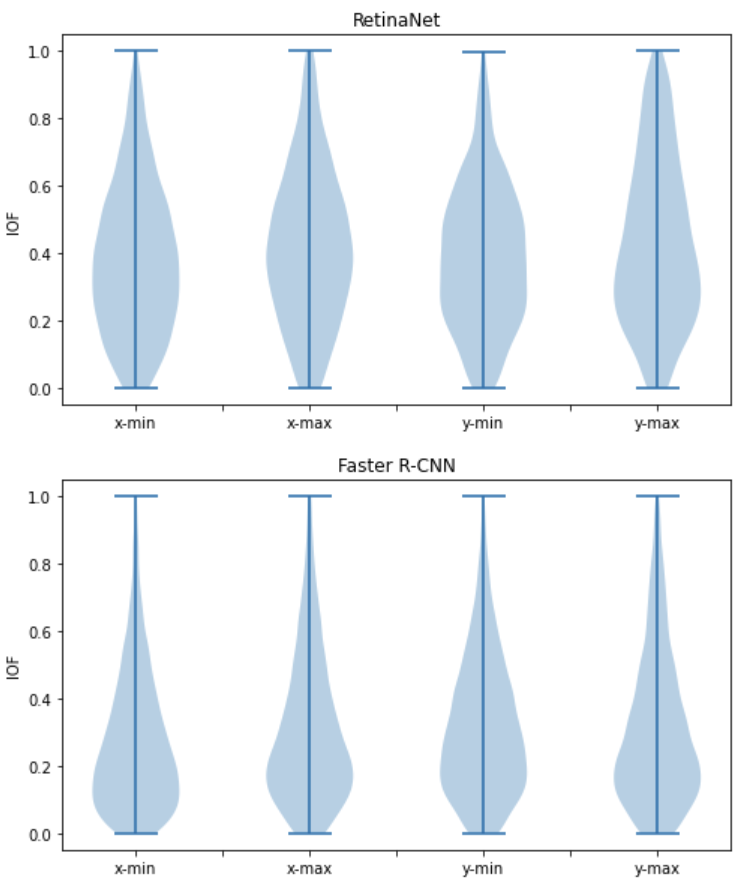}
  \caption{Violin plots showing the IOF values per detected object for RetinaNet and Faster R-CNN.}
  \label{fig:violin}
\end{figure}

\subsection{Validation}

To validate our method, we performed tests using the MSCOCO dataset for a random set of classes. We began by generating saliency maps using the SmoothGrad algorithm with the Detectron2 library \cite{wu2019}. Separate sets of saliency maps were generated for RetinaNet and Faster-RCNN to show viability using one-stage and two-stage object detectors. We also restrict the sample to high confidence detects, where the classification score is above 0.9. Next, for the purposes of validation, we created a binarized segmentation mask using the saliency map. To accomplish this, we used a very simplistic algorithm, where we first applied a 2$\sigma$ Gaussian filter to smooth out the map, and then used pixels that were greater than a factor of 0.32 of the max pixel value as the foreground, while setting the remaining pixels to zero.

The next step involved using the ground truth segmentation polygons from MSCOCO. Direct comparisons against these polygons did not adequately demonstrate the capabilities of our technique, since the goal is to show that generating a saliency map for the bounding box parameters (such as x-min) would show greater significance around the corresponding parameter (the left side of the ground truth for x-min). As a result, we performed the following method of evaluating performance of the saliency map with the corresponding bounding box parameter:
\begin{itemize}
  \item{Divide the ground truth segmentation polygons into thirds in the x and y dimensions (Figure \ref{fig:val_a})}
  \item{Calculate the intersection over foreground (IOF) of the first and last thirds of the ground truth polygon with the corresponding saliency masks from the min and max bounding box parameters (Figure \ref{fig:val_b}).}
  \item{Calculate the IOF of the saliency mask with the middle section of the ground truth polygon to determine background.}
\end{itemize}

Comparing the saliency mask with the middle section is to demonstrate that the mask for the bounding box parameters do tend to be localized on the respective sides, rather than spreading out into other parts of the ground truth segmentation polygon.

The use of IOF, with the saliency segmentation mask area as the denominator, was used over Dice or Jaccard metrics to more appropriately capture relevance of the intersection. Dividing the ground truth polygon into thirds in the x and y dimensions provides an automated and consistent approach, but results in variations based on the size and orientation of the objects that skew the values of the aforementioned metrics.

\section{Conclusions}

The use of saliency methods is a popular way of achieving explainability of a model, and the extension of these methods into object detection algorithms provides visibility into their predictions. Applying these methods to both the class prediction as well as the bounding box parameters allows confirmation that the results are based on the relevant features. The results from Figure \ref{fig:violin} demonstrate a significant overlap between the generated saliency masks and the localized sections of the ground truth polygons. Some of the low values can be attributed to the imperfections in the ground truth segmentation of the objects, whereas others are the result of the saliency map picking up features in the image that do not directly correspond with the detected class.

\section{Acknowledgements}

We thank Dr. Tim Oates (UMBC) for review and comments on manuscript. This work was performed for CalypsoAI (https://calypsoai.com).

{\small
\bibliographystyle{ieee_fullname}
\bibliography{egbib}
}

\end{document}